\documentclass{article}

\usepackage{microtype}
\usepackage{graphicx}
\usepackage{subfigure}
\usepackage{booktabs} 

\usepackage{hyperref}

\usepackage[accepted]{icml2020}

\usepackage{subfigure}
\usepackage{bm}
\usepackage{soul}
\usepackage{diagbox}
\usepackage{multirow}
\usepackage{amsmath}
\usepackage{amsfonts}
\usepackage{colortbl}
\usepackage{url}
\usepackage{subfigure}
\usepackage{colortbl}
\usepackage{verbatim}
\usepackage{bm}
\usepackage{soul}
\usepackage{bbm}
\usepackage{bigstrut,bigdelim}
\usepackage{paralist}
\usepackage{algorithm}
\usepackage{algorithmic}
\usepackage{makecell}
\usepackage{diagbox}
\usepackage{multirow}
\usepackage{float}
\usepackage{mathrsfs}
\usepackage{textcomp}
\usepackage{color}
\usepackage[switch]{lineno}
\usepackage{authblk}

\newcommand{\thickhline}{\noalign{\hrule height 1pt}}
\newcommand\Tstrut{\rule{0pt}{2ex}}         

\begin{document}

\twocolumn[
\icmltitle{Learning an Effective Context-Response Matching Model \\ with Self-Supervised Tasks for Retrieval-based Dialogues}

\icmlsetsymbol{equal}{*}

\begin{icmlauthorlist}
\icmlauthor{Ruijian Xu }{equal,pku}
\icmlauthor{Chongyang Tao}{equal,micro}
\icmlauthor{Daxin Jiang}{micro}
\icmlauthor{Xueliang Zhao}{pku_2}
\icmlauthor{Dongyan Zhao}{pku}
\icmlauthor{Rui Yan}{pku}
\end{icmlauthorlist}

\icmlaffiliation{pku}{Wangxuan Institute of Computer Technology, Peking University, Beijing, China}
\icmlaffiliation{micro}{Microsoft Corporation, Beijing}
\icmlaffiliation{pku_2}{Center for Data Science, Peking University, Beijing, China}

\icmlcorrespondingauthor{Rui Yan}{ruiyan@pku.edu.cn}
\icmlcorrespondingauthor{Chongyang Tao}{chongyangtao@gmail.com}

\vskip 0.3in
]



\printAffiliationsAndNotice{\icmlEqualContribution} 

\begin{abstract}
Building an intelligent dialogue system with the ability to select a proper response according to a multi-turn context is a great challenging task.
Existing studies focus on building a context-response matching model with various neural architectures or PLMs and typically learning with a single response prediction task.
These approaches overlook many potential training signals contained in dialogue data, which might be beneficial for context understanding and produce better features for response prediction. 
Besides, the response retrieved from existing dialogue systems supervised by the conventional way
still faces some critical challenges, including incoherence and inconsistency.
To address these issues, in this paper, we propose learning a context-response matching model with auxiliary self-supervised tasks designed for the dialogue data based on pre-trained language models.
Specifically, we introduce four self-supervised tasks including next session prediction, utterance restoration, incoherence detection and consistency discrimination, and jointly train the PLM-based response selection model with these auxiliary tasks in a multi-task manner.  
By this means, the auxiliary tasks can guide the learning of the matching model to achieve a better local optimum and select a more proper response.
Experiment results on two benchmarks indicate that the proposed auxiliary self-supervised tasks bring significant improvement for multi-turn response selection in retrieval-based dialogues, and our model achieves new state-of-the-art results on both datasets.

\end{abstract}

\section{Introduction}

Building a dialogue system that can converse with people naturally and meaningfully is one of the most challenging problems towards high-level artificial intelligence, and has been drawing increasing interests from both academia and industry area.
Most existing dialogue systems are either generation-based~\cite{vinyals2015neural,serban2016building} or retrieval-based~\cite{wang-etal-2013-dataset,lowe-etal-2015-ubuntu,wu-etal-2017-sequential,tao2019multi}.
Given the dialogue context, generation-based approaches synthesize a response word by word with a conditional language model, while retrieval-based methods select a proper response from a candidate pool.
In this paper, we focus on retrieval-based approaches that are superior in providing informative responses and have been widely applied in several famous commercial products such as XiaoIce~\cite{shum2018eliza} from Microsoft and AliMe Assist~\cite{li2017alime} from Alibaba.

We consider the response selection task in multi-turn dialogues, where the retrieval model ought to select a most proper response by measuring the matching degree between a multi-turn dialogue context and a number of response candidates.
Earlier studies~\cite{wang-etal-2013-dataset,hu2014convolutional,lowe-etal-2015-ubuntu} concatenate the context to a single utterance and calculate the matching score with the utterance-level representations.
Later, most response selection models \cite{zhou-etal-2016-multi,wu-etal-2017-sequential,zhang-etal-2018-modeling} perform context-response matching within the representation-matching-aggregation paradigm, where each turn of utterance is represented individually and sequential information is aggregated among a sequence of utterance-response matching features.
To further improve the performance of response selection,
some recent approaches consider multiple granularities (or layers) of representations~\cite{zhou2018multi,tao2019multi,wang2019multi} for matching or propose more complicated interaction mechanisms between the context and the response~\cite{tao-etal-2019-one}.

Recently, a wide range of studies have shown that pre-trained language models (PLMs), such as BERT~\cite{devlin-etal-2019-bert}, XLNET~\cite{yang2019xlnet} and RoBERTa~\cite{liu2019roberta}, on the large corpus can learn universal language representations, which are helpful for various downstream natural language processing tasks and can get rid of training a new model from scratch.
To adapt pre-trained models for multi-turn response selection, \citet{whang2020domain} and \citet{gu2020speaker} make the first attempt to  utilize BERT~\cite{devlin-etal-2019-bert} to learn a matching model, where context and the candidate response are first concatenated and then fed into the PLMs for calculating the final matching score. 
These pre-trained language models can well capture the interaction information among inter-utterance and intra-utterance through multiple transformer layers.
Although PLM-based response selection models demonstrate superior performance due to its strong representation ability, it is still challenging to effectively learn task-related knowledge during the training process, especially when the size of training corpora is limited.
Naturally, these studies typically 
learn the response selection model with only the context-response matching task
and overlook many potential training signals 
contained in dialogue data. 
Such training signals might  be  beneficial  for  context  understanding  and  produce better  features  for  response  prediction. 
Besides, the response retrieved by existing dialogue systems supervised by the conventional way still faces some critical challenges, including  incoherence  and  inconsistency.

On account of the above issues, in this paper, instead of configuring complex context-response matching models,
we propose learning the context-response matching model with auxiliary self-supervised tasks designed for dialogue data based on pre-trained language models (e.g., BERT).
Specifically, we introduce four self-supervised tasks  including  \emph{next session prediction}, \emph{utterance restoration}, \emph{incoherence  detection} and \emph{consistency  discrimination}, and  jointly  train  the  PLM-based  response  selection  model with  these  auxiliary  tasks  in  a  multi-task  manner. 
On the one hand, these auxiliary tasks help improve the capability of the response selection model to understand the dialogue context and measure the semantic relevance, consistency or coherent between the context and the response candidates. On the other hand, they can guide the matching model to effectively learn task-related knowledge with a fixed amount of train corpora and produce better features for response prediction. 

We conduct experiments on two benchmark data sets for multi-turn response selection: the Ubuntu Dialog Corpus~\cite{lowe-etal-2015-ubuntu} and the E-commerce Dialogue Corpus~\cite{zhang-etal-2018-modeling}.
Evaluation results show that our proposed approach is significantly better than all state-of-the-art models on both datasets.
Compared with the previous state-of-the-art methods, our model achieves 2.9\% absolute improvement in terms of $R_{10}@1$ for the Ubuntu dataset and 4.8\% absolute improvement for the E-commerce dataset.
Furthermore, we applied our proposed self-supervised learning schema to some non-PLM-based response selection models, e.g., dual LSTM~\cite{lowe-etal-2015-ubuntu} and ESIM~\cite{chen2019sequential}.
Experimental results indicate that our learning schema can also bring consistent and significant improvement to the performance of the existing matching models. Surprisingly, with self-supervised learning, a simple ESIM even performs better than BERT on the ubuntu dataset, demonstrating that our approach is beneficial for various matching architectures.

In summary, our contributions are three-fold:
\begin{itemize}
\item We propose learning a context-response matching model with multiple auxiliary self-supervised tasks to fully utilize various training signals in the multi-turn dialogue context.
\item We design four self-supervised tasks, aiming at enhancing the capability of a PLM-based response prediction model in capturing the semantic relevance, coherence or consistency. 
\item We  achieve new state-of-the-art results on two benchmark datasets. Besides, with the help of auxiliary self-supervised tasks, a simple ESIM model can even achieve better performance than BERT on the Ubuntu dataset.
\end{itemize}

\section{Model}
\label{approach}

\subsection{Task Formalization}

Suppose that there is a multi-turn dialogue dataset $\mathcal{D} = \{c_i, r_i, y_i\}_{i=1}^{N}$, where $c_i=\{u_{i,1}, u_{i,2}, \ldots, u_{i,m_i}\}$ denotes a dialogue context with $u_{i,t}$ representing the utterance of the $t$-th turn, $r_i$ denotes a response candidate, and $y_i \in \{0,1\}$ denotes a label with $y_i=1$ indicating that $r_i$ is a proper response for $c_i$ (otherwise, $y_i=0$).
The task is to learn a matching model $g(\cdot,\cdot)$ from $\mathcal{D}$ so that for any new context $c=\{u_1,u_2,\ldots,u_m\}$ and a response candidate $r$, $g(c,r) \in [0,1]$ can measure the matching degree between $c$ and $r$.

\begin{figure}[t!]
  \centering
  \includegraphics[width=1.01\linewidth]{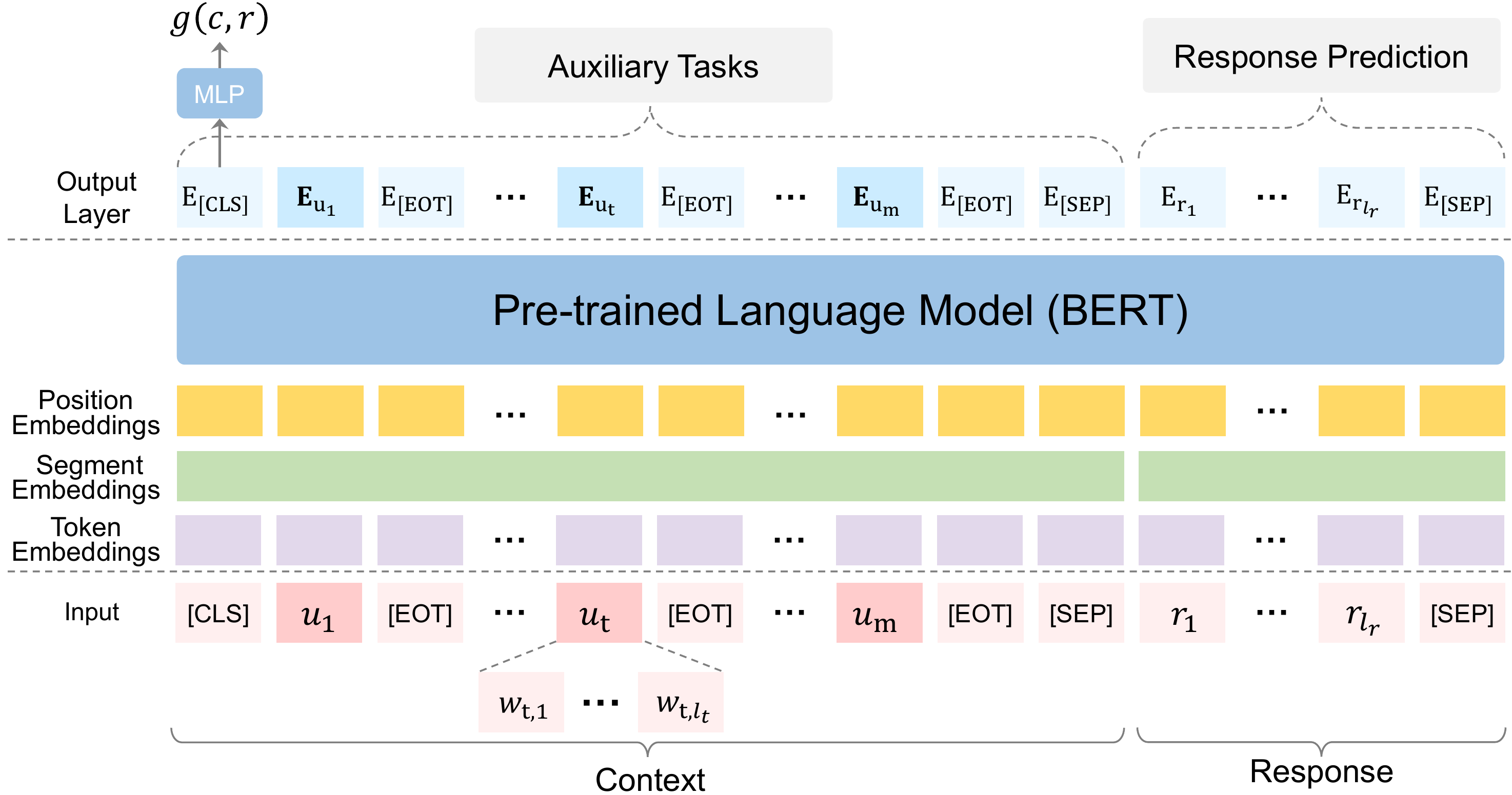}
  \vspace{-3mm}
  \caption{Overall architecture of our model.}
  \vspace{-3mm}
  \label{fig:main-architecture}
\end{figure}

\subsection{Matching with PLMs}
We consider building the context-response matching model with the pre-trained language models, as it is trained on large amounts of unlabelled data and provides strong “universal representations" that can be finetuned on task-specific training data to achieve good performance on downstream tasks.
Following previous studies \cite{gu2020speaker,whang2020domain}, we select BERT as the base model for a fair comparison.

Specifically, given a context $c = \{u_1,u_2,\ldots,u_m\}$, where the $t$-th utterance $u_t=\{w_{t,1},\ldots,w_{t,l_t}\}$  is a sequence with $l_t$ words, a response candidate $r = \{r_1,r_2,\ldots,r_{l_r}\}$ consisting of $l_r$ words and a label $y\in\{0, 1\}$,
we first concatenate all utterances in the context and the response candidate as a single consecutive token sequence with special tokens separating them, which can be formulated as
$x = \{  [\texttt{CLS}], u_1, [\texttt{EOT}], u_2,  [\texttt{EOT}], \ldots,  [\texttt{EOT}], u_m, $ $ [\texttt{EOT}],  [\texttt{SEP}], r, [\texttt{SEP}] \}$.
Here $[\texttt{CLS}]$ and $[\texttt{SEP}]$ are the classification symbol and the segment separation symbol of BERT, $[\texttt{EOT}]$ is the "End Of Turn" tag designed for multi-turn context.
For each word of $x$, \emph{token}, \emph{position} and \emph{segment} embeddings of $x$ are summated and fed into pre-trained transformer layer (a.k.a. BERT), giving us the contextualized embedding sequence $ \{E_{\texttt{[CLS]}}, E_2, \ldots, E_{l_x}\}$. 
$E_{\texttt{[CLS]}}$ is an aggregated representation vector that contains the semantic interaction information for the context-response pair.
We then fed $E_{\texttt{[CLS]}}$ into a multi-perception layer to obtain the final matching score for the context-response pair:
\begin{equation} 
\label{eq-rm}
\begin{aligned} 
g(c,r) &= \sigma ({W}_{2} \cdot f({W}_1 E_{\texttt{[CLS]}} + b_1) + b_{2})
\end{aligned}
\end{equation}
where ${W}_{\{1,2\}}$ and $b_{\{1,2\}}$ are trainable parameters for response prediction task, $f(\cdot)$ is a $\mathtt{tanh}$ activation function, $\sigma(\cdot)$ stands a sigmoid function.

Finally, cross-entropy loss function is utilized as the training objective of the \emph{context-response matching} task:
\begin{equation} 
\label{loss-rm}
\begin{aligned} 
\mathcal{L}_\texttt{crm} &=  -y \log(g(c,r)) - (1-y)\log(1-g(c,r))
\end{aligned}
\end{equation}

Before the fine-tuning procedure with the above context-response matching task,
for a fair comparison, we follow previous studies~\cite{whang2020domain,gu2020speaker,gururangan-etal-2020-dont} and carry out domain-adaptive post-training to incorporate in-domain knowledge into BERT.
In the rest of this section, we will introduce our proposed four auxiliary self-supervised tasks, and then present the final learning objective of our model.

\subsection{Self-Supervised Tasks}

\begin{figure}[t!]
  \centering
   \includegraphics[width=\linewidth]{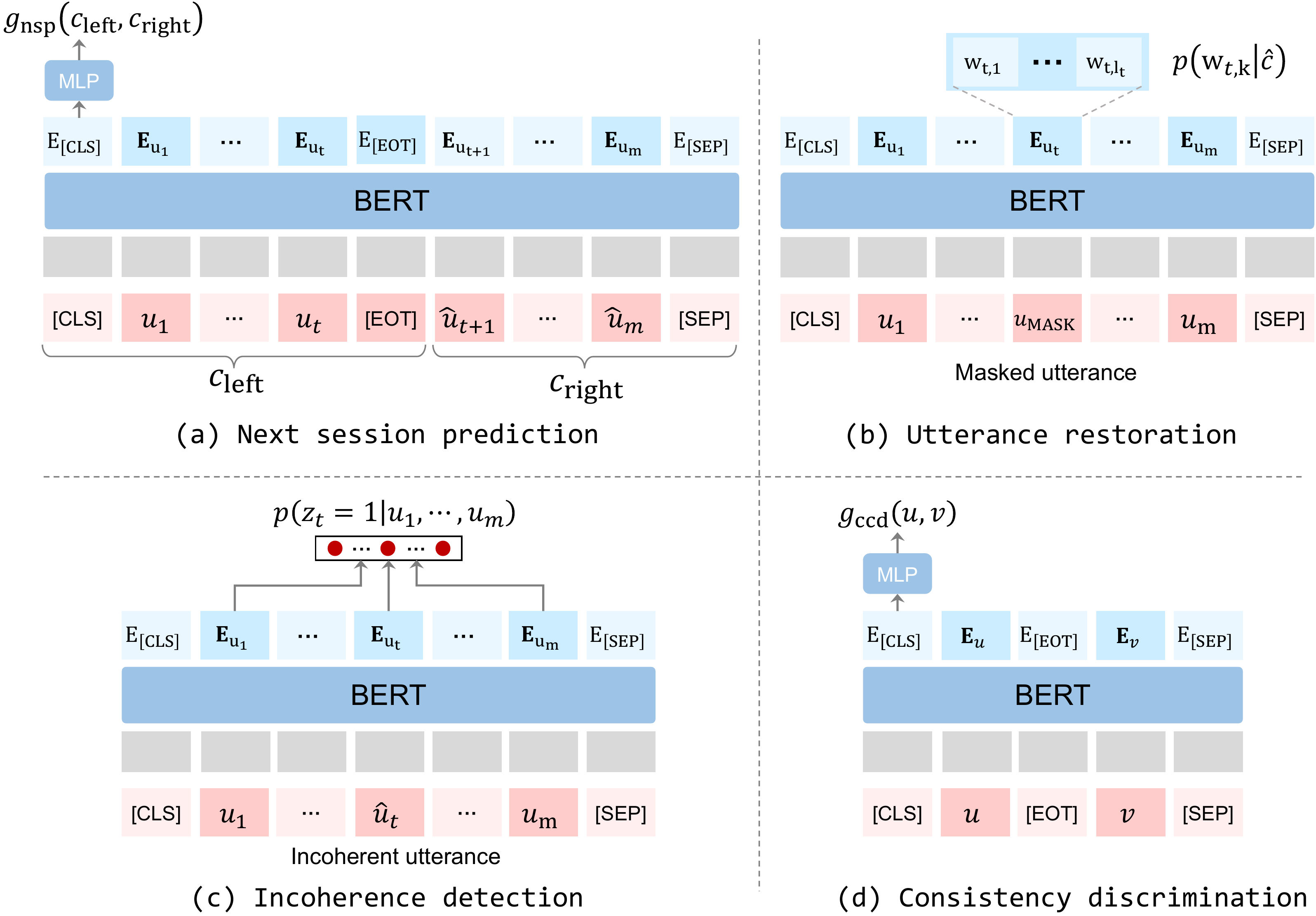}
  \caption{Sketches of four types of self-supervised tasks. Gray square stands for various embeddings for each token.}
  \label{fig:tasks}
\end{figure}

Heading for a matching model that can effectively learn domain knowledge with a fixed amount of training corpora and produce better features for response prediction,
we design four auxiliary self-supervised tasks,
i.e. \emph{session-level matching}, \emph{utterance restoration}, \emph{incoherence detection} and \emph{ consistency classification}. These self-supervised tasks try to enhance the capability of the model to measure the semantic relevance, coherent, and  consistency between the context and the response candidate.
On the other hand, they can also  guide the learning of the model to achieve a better local optimum. Figure~\ref{fig:tasks} illustrates the sketches of four types of self-supervised tasks.

\subsubsection{Next Session Prediction}
Due to the natural sequential relationship between dialogue turns, the latter turns usually show a strong semantic relevance with the previous turns in the context. Inspired by such a characteristic, we design a more general response prediction task with the dialogue context, name \emph{next session prediction} (NSP), to fully utilize the sequential relationship of the dialogue data and enhance the capability of the model to measure the semantic relevance.
Specifically, the next session prediction task requires the model to predict whether two sequences are consecutive and relevant.
However, instead of matching a context with a response utterance, the model needs to calculate the matching degree between two pieces of dialogue session.

Formally, given a context $c=\{u_1,u_2,\ldots,u_m\}$, we randomly\footnote{In this work, all random sampling operations are carried out according to uniformly distribution.} split $c$ into two consecutive pieces $c_\texttt{left} = \{u_1,\ldots,u_t\}$ and $c_\texttt{right} = \{u_{t+1},\ldots,u_{m}\}$.
Then, with a 50\% chance, we replace $c_\texttt{left}$ or $c_\texttt{right}$ with a piece of context
sampled from the whole training corpus\footnote{If $c_\texttt{left}$ is replaced, the new piece should be the left part of another context with a random length, and vice versa.}.
If one of the two piece is replaced, we set the label $y_\texttt{nsp} = 0$, otherwise $y_\texttt{nsp} = 1$.
The next session prediction task requires the model to discriminate whether $c_\texttt{left}$ and $c_\texttt{right}$ can form a consecutive context.

To train PLMs with the proposed self-supervised task, we first concatenate all utterances of each piece as a single sequence with \texttt{[EOT]} appended to the end of each utterance. 
Similar to the main task, we fed two segments into BERT encoder and obtain the aggregated representation of the piece pair $E_{\texttt{[CLS]}}^\texttt{nsp}$.
We further compute the final matching score $g_\texttt{nsp}(c_\texttt{left},c_\texttt{right})$ with a non-linear transformation.
Finally, the objective function of context alignment task can be formulated as
\begin{equation} 
\label{nsp-loss}
\begin{aligned} 
\mathcal{L}_\texttt{nsp} &=  -y_\texttt{nsp} \log(g_\texttt{nsp}(c_\texttt{left},c_\texttt{right})) \\
&- (1-y_\texttt{nsp})\log(1-g_\texttt{nsp}(c_\texttt{left},c_\texttt{right}))
\end{aligned}
\end{equation}

\subsubsection{Utterance Restoration}
As one of the common self-supervised tasks in PLMs, token-level masked language modeling is usually utilized to guide the model to learn semantic and syntactic features of word sequences with the bidirectional context.
Here we further introduce utterance-level masked language modeling, i.e. \emph{utterance restoration} (UR) task to encourage the model to be aware of the semantic connections among utterances in the context.
Specifically, we mask all the tokens in an utterance  randomly sampled from the dialogue session and let the model restore it with the information from the rest context.
By learning to predict a proper utterance that fits its surrounding dialogue context, the model can produce better representations that can well adapt to dialogues, similar to the idea of continuous bag-of-words model~\cite{mikolov2013efficient}.

Formally, given a context $c=\{u_1,u_2,\ldots,u_m\}$, we randomly select an utterance $u_t$ and replace all tokens in the utterance with a special token \texttt{[MASK]}.
The model is required to restore $u_t$ based on $\hat{c} = \{u_1, \ldots, u_{t-1}, u_{\text{mask}}, u_{t+1},\ldots, u_m\}$.
To adapt the task in BERT, we formulate input of BERT encoder as $x_\texttt{ur} = \{\texttt{[CLS]}, u_1,\texttt{[EOT]}, \ldots, u_\texttt{mask}, \texttt{[EOT]}, \ldots, u_m,  \texttt{[EOT]}, \\ \texttt{[SEP]}\}$, where $u_\texttt{mask}$ consists of only \texttt{[MASK]} tokens and has the same length with $u_t$.
After being processed by BERT, the top layer output a representation sequence $E_\texttt{ur} = \{E_{\texttt{[CLS]}}, E_{1,1}, \ldots, E_{1,l_1}, E_{\texttt{[EOT]}}, \ldots, E_{m,1}, \ldots, E_{m,l_m}, \\ E_{\texttt{[EOT]}},E_{\texttt{[SEP]}}\}$, where $l_t$ is the length of the $t$-th utterance. The model will predict the masked utterance conditioned on the contextualized representations of each word. The probability distribution of each masked word can be calculated as
\begin{equation} 
\begin{aligned} 
E_{t,j}^\prime &= \text{GLEU}({W}_{ur} E_{t,j} + b_{ur}) \\
p(w_{t,j}|\hat{c}) &= \text{softmax}\big({W}_{ur}^{\prime} E_{t,j}^\prime  + b_{ur}^{\prime} \big)
\end{aligned}
\end{equation}
where ${W}_\texttt{ur}$, ${W}_\texttt{ur}^{ \prime }$, $b_\texttt{ur}$, $b_\texttt{ur}^{ \prime }$ are trainable parameters, $w_{t,j}$ is the $j$-th token of the $t$-th utterance, and $\text{GLEU}(\cdot)$ is an activation function.
Then, the training objective of \emph{utterance restoration task} is to minimize the following negative log-likelihood (NLL):
\begin{equation} 
\begin{aligned} 
 \mathcal{L}_\texttt{ur} &= - \frac{1}{l_t}\sum_{j=1}^{l_t}\log p(w_{t,j}|\hat{c}) \\
\end{aligned}
\end{equation}
\subsubsection{Incoherence Detection}
Inspired by the concept of discourse coherence \cite{jurafsky2000speech} in linguistics, we further introduce the \emph{incoherence detection} (ID) task which requires the model  to  recognize  the  incoherent  utterance  within  a  dialogue session, so as to enhance the capability of a model on capturing the sequential relationship among utterances and selecting coherent response candidates.
Specifically, given a dialogue context $c=\{u_1,\ldots,u_m\}$, we randomly select one of the utterances $u_k \in \{u_1, \ldots, u_m\}$ and replace it with an utterance randomly sampled from the whole training corpus.
Then, the model should find the incoherent utterance among the context. For each sample, we define a one-hot label $\{z_1, \ldots, z_m\}$ , where $z_t = 1$ if $t=k$, indicating that the $t$-th utterance is been replaced, otherwise $z_t=0$. 

To model this task, BERT encoder takes an input $x_\texttt{id} = \{\texttt{[CLS]}, u_1,  \texttt{[EOT]}, \ldots, u_m, \texttt{[EOT]}, \texttt{[SEP]}\}$
and outputs $E_\texttt{id} = \{E_{\texttt{[EOT]}}, E_{1,1},  \ldots, E_{m,l_m}, E_{\texttt{[SEP]}}\}$,
where $E_{t,j}$ denotes the contextualized embedding of the $j$-th word in the $k$-th utterance and $l_t$ is the length of $t$-th utterance.
We calculate the aggregated representation of the $k$-th utterance by fusing the mean and max value of the embedding sequence $\{E_{t,1}, \ldots, E_{t,l_t}\}$, which can be formulated as
\begin{equation} 
\begin{aligned} 
  U_t &= \big[ \frac{1}{l_t}\sum_{j=1}^{l_t}E_{t,j}; \max \limits_{1 \le j \le l_t} E_{t,j} \big]
\end{aligned}
\end{equation}
Then, the model makes a prediction based on the aggregated  representations of each utterance, the probability of the $t$-th utterance being replaced is
\begin{equation} 
\begin{aligned} 
 p(z_t=1|u_1, \ldots, u_m) &=  \text{softmax}({W}_\texttt{id}U_t + b_\texttt{id}) \\
  &= \frac{\exp({W}_\texttt{id}U_t + b_\texttt{id})}{\sum_{s=1}^{m} \big( \exp({W}_\texttt{id}U_s)+ b_\texttt{id} \big)}
\end{aligned}
\end{equation}
where ${W}_\texttt{id}$ and $b_\texttt{id}$ are trainable parameters.

Finally, the learning objective of inconsistency detection task is defined as
\begin{equation} 
\begin{aligned} 
 \mathcal{L}_\texttt{id} &= - \sum_{t=1}^{m} z_t \log p(z_t=1|u_1, \ldots, u_m)
\end{aligned}
\end{equation}

\subsubsection{Consistency Discrimination} 
Selecting responses that are consistent with the dialogue context is one of the major challenges
in building engaging conversational agents. However, most previous studies focused on modeling the semantic relevance between the context and the response candidate.
Intuitively, utterances from the same dialogue session tend to share similar topics, and utterances from the same interlocutor tend to share the same personality or style.
According to the characteristics, we attempt to enhance the capability of a response prediction model to measure the  consistency with a self-supervised discriminative training scheme that utilizes the natural structure of dialogue data.

Formally, given a dialog context $c=\{u_1,u_2,\ldots,u_m\}$, we sample two utterances from the same interlocutor\footnote{We assume that utterances in a dialogue context are posed one by one, therefore we can simply sample utterances from only the odd turns or even turns.}, and denote them as $u$ and $v$ respectively.
Then, we randomly sample an utterance $\tilde{v}$ from another context in the training corpus.
The model is required to measure the consistency degree of  $\langle u, v\rangle$ and  $\langle u, \tilde{v} \rangle$ and give a higher score to $\langle u, v \rangle$.
Since $u$ and $v$ are not consecutive in the dialogue context and from the same interlocutor, the model is encouraged to capture the features about the consistency (such as topic, personality and style) between two sequences, rather than semantic relevance or coherence.

To calculate the consistency score of a sequence pair $\langle u, v \rangle$, we first concatenate the two utterances
as 
$x_\texttt{cd} = \{\texttt{[CLS]}, u, \texttt{[SEP]},  v, \texttt{[SEP]}\}$, and then fed the sequence into BERT.
As described in previous tasks, BERT returns an aggregated representation $E_\texttt{[CLS]}^\texttt{cd}$.
Then, the consistency score
$g_\texttt{cd}(u,v)$ is computed with a non-linear transformation over $E_\texttt{[CLS]}^\texttt{cd}$.
Likewise, we can obtain the consistency score of $\langle u, \tilde{v} \rangle$, i.e. $g_\texttt{cd}(u,\tilde{v})$.
Finally, we would like $g_\texttt{cd}(u,v)$ to be larger than $g_\texttt{cd}(u,\tilde{v})$ by at least a margin $\Delta$ and define
the learning objective as a hing loss function:
\begin{equation} 
\label{eq:cdloss}
\begin{aligned}
\mathcal{L}_\texttt{cd} &= \max \{ 0, \Delta -  g_\texttt{cd}(u,v) + g_\texttt{cd}(u,\tilde{v})\}
\end{aligned}
\end{equation}

\subsection{Learning Objective}
We adopt a multi-task learning manner and define the final objective function as:
\begin{equation}  \label{eq:loss}
\begin{aligned} 
\mathcal{L}_\texttt{final} &= \mathcal{L}_\texttt{crm} + \alpha \mathcal{L}_\texttt{self} \\
 \mathcal{L}_\texttt{self} &= \mathcal{L}_{\texttt{slm}} +  \mathcal{L}_{\texttt{ur}} +  \mathcal{L}_{\texttt{id}} +  \mathcal{L}_{\texttt{cd}} \\
\end{aligned}
\end{equation}
where $\alpha$ is a hyper-parameter as
a trade-off between the objective of the main task and those of the auxiliary tasks.
In this way, all tasks are joint learned so that the model can effectively leverage the training corpus and learn both characteristics of dialogue text and implicit knowledge contained in the dialogue data.
The auxiliary tasks can be regarded as regularization in model estimation for enhancing the model's generalization ability. 
  \begin{table*}[ht!]
    \centering
    \resizebox{0.85\textwidth}{!}{
    \begin{tabular}{c|l|c|c|c|c|c|c|c}
      \thickhline 
      \multicolumn{2}{c|}{ \multirow{2}{*}{\backslashbox{Models}{Metrics}}} &  
      \multicolumn{4}{c|}{\textbf{Ubuntu Corpus}} &
      \multicolumn{3}{c}{\textbf{E-commerce Corpus}}
      \\ \cline{3-9}
      \multicolumn{2}{c|}{} &  R$_2$@1 &  R$_{10}$@1 & R$_{10}$@2 & R$_{10}$@5 
      & R$_{10}$@1 & R$_{10}$@2  & R$_{10}$@5 \\ \hline
      &  DualLSTM~\cite{lowe-etal-2015-ubuntu} & 0.901 & 0.638 & 0.784 & 0.949 & 0.365 & 0.536 & 0.828  \\ 
      &  Multi-View~\cite{zhou-etal-2016-multi}  & 0.908 & 0.662 & 0.801 & 0.951 & 0.421 & 0.601 & 0.861 \\ 
       & SMN~\cite{wu-etal-2017-sequential} & 0.926 & 0.726 & 0.847 & 0.961 & 0.453 & 0.654 & 0.886 \\
       & DUA~\cite{zhang-etal-2018-modeling} &- & 0.752 & 0.868 & 0.962 & 0.501 & 0.700 & 0.921\\
       Non-PLM-based & DAM~\cite{zhou2018multi} & 0.938 & 0.767 & 0.874 & 0.969 &  0.526 & 0.727 & 0.933  \\
       Models & MRFN~\cite{tao2019multi} & 0.945 & 0.786 & 0.886 & 0.976 & - & - & - \\
       & IMN~\cite{gu2019interactive} & 0.946 & 0.794 & 0.889 & 0.974 & 0.621 & 0.797 & 0.964 \\
       & ESIM~\cite{chen2019sequential} & 0.950 & 0.796 & 0.874 & 0.975 & 0.570 & 0.767 & 0.948\\
       & IoI~\cite{tao-etal-2019-one} &  0.947 & 0.796 & 0.894 & 0.974 & 0.563 & 0.768 & 0.950 \\
       & MSN~\cite{yuan-etal-2019-multi} & -  & 0.800 & 0.899 & 0.978 & 0.606 & 0.770 & 0.937 \\\hline
       & BERT~\cite{whang2020domain} & 0.954 & 0.817 & 0.904 & 0.977 & 0.610 & 0.814 & 0.973 \\ 
      & SA-BERT \cite{gu2020speaker} & 0.965 & 0.855 & 0.928 & 0.983 & 0.704 & 0.879 & 0.985 \\ 
      & BERT-VFT~\cite{whang2020domain} & - & 0.855 & 0.928 & 0.985 & - & - & - \\
      & BERT-VFT (Ours)\Tstrut & 0.969 & 0.867 & 0.939 & 0.987 &  0.717 & 0.884 & 0.986\\\cline{2-9}
      PLM-based & {BERT-SL} \Tstrut & \textbf{0.975}* & \textbf{0.884}* & \textbf{0.946}* & \textbf{0.990}* & \textbf{0.776}* & \textbf{0.919}* & 0.991\\ \cline{2-9}
      Models & {BERT-SL} w/o.  NSP \Tstrut& 0.973 & 0.879 & 0.944 & 0.989 & 0.760 & 0.914 & 0.988 \\
      & {BERT-SL} w/o. UR & 0.974 & 0.881 & 0.945 & 0.990 & 0.763 & 0.916 & 0.991 \\
      & {BERT-SL} w/o. ID & 0.972 & 0.877 & 0.942 & 0.989 & 0.755 & 0.911 & 0.987 \\
      & {BERT-SL} w/o. CD & 0.973 & 0.880 & 0.945 & 0.989 & 0.742 & 0.897 & 0.986 \\ 
      \thickhline 
    \end{tabular}
    }
    \caption{Evaluation results on the two data sets.   Numbers marked with $*$ mean that the improvement is statistically significant compared with the baseline (t-test with $p$-value $<0.05$). Numbers in bold indicate the best strategies for the corresponding models on specific metrics.} 
    \label{exp:main-results}
  \end{table*}
\section{Experiments}
\subsection{Datasets and Evaluation Metrics}
we evaluate the proposed method on two benchmark datasets for multi-turn dialogue response selection.
The first dataset is the \emph{\textbf{Ubuntu Dialogue Corpus (v1.0)}}~\cite{lowe-etal-2015-ubuntu}, which consists of multi-turn English dialogues about technical support and is collected from chat logs of the Ubuntu forum.
We use the copy shared by \citet{gu2020speaker}, in which numbers, paths and URLs are replaced by placeholders.
The Ubuntu dataset contains $1$ million context-response pairs for training, and $0.5$ million pairs for validation and test.

The ratio of positive candidates and negative candidates is $1:1$ in the training set, and $1:9$ in the validation set and the test set. The second dataset is the \emph{\textbf{E-commerce Dialogue Corpus}} ~\cite{zhang-etal-2018-modeling},
which consists of real-world multi-turn dialogues between customers and customer service staff on Taobao\footnote{\url{https://www.taobao.com}}, the largest e-commerce platform in China.
The E-commerce dataset contains $1$ million context-response pairs for training, and $10$ thousand pairs for validation and test.
The ratio of positive candidates and negative candidates is $1:1$ in the training set and the validation set, and $1:9$ in the test set.

Following \citet{lowe-etal-2015-ubuntu} and \citet{zhang-etal-2018-modeling}, we employ $R_n@k$s as evaluation metrics, where $R_n@k$ denotes recall at position $k$ in $n$ candidates and measures the probability of the positive response being ranked in top $k$ positions among $n$ candidates.

\subsection{Baseline Models}
We compared BERT-SL with the following models:

\textbf{DualLSTM}~\cite{lowe-etal-2015-ubuntu}: 
 the model concatenates all utterances in the context to form a single sequence and calculates a matching score  based on the representations produced by an LSTM.
 
\textbf{Multi-View}~\cite{zhou-etal-2016-multi}:
 the model measures the matching degree between the context and the response candidate in both a word view and an utterance view.

\textbf{SMN}~\cite{wu-etal-2017-sequential}:
 the model lets each utterance in the context interacts with the response candidate, and the matching vectors of all utterance-response pairs are aggregated with an RNN to calculate a final matching score. 
 
\textbf{DUA}~\cite{zhang-etal-2018-modeling}: the model formulates previous utterances into context using a deep utterance aggregation model, and performs context-response similar to SMN.

\textbf{DAM}~\cite{zhou2018multi}: 
 the model is similar to SMN, but utterances in the context and the response candidate are represented with stacked self-attention and cross-attention layers. The matching vectors are aggregated with a 3-D CNN. 

\textbf{MRFN}~\cite{tao2019multi}:
 the model employs multiple types of representations for context-response interaction, where each type encodes semantics of units from a kind of granularity or dependency among the units.
 
\textbf{ESIM}~\cite{chen2019sequential}: the model first concatenates all utterances in the context into a single sequence, and then employs ESIM structure derived from NLI for context-response matching.

\textbf{IMN}~\cite{gu2019interactive}: following ~\citet{wu-etal-2017-sequential}, the model enhances the representations at both the word- and sentence-level and collects matching information of utterance-response pairs bidirectionally.

\textbf{IoI}~\cite{tao-etal-2019-one}:
the model lets the context-response matching process goes deep along the interaction block chain via representations in an iterative fashion.

\textbf{MSN}~\cite{yuan-etal-2019-multi}:
 the model utilizes a multi-hop selector to select the relevant utterances in context and then matches the filtered context with the response candidate to obtain a matching score.

\textbf{BERT}~\cite{whang2020domain}:
the model fine-tunes the BERT with the concatenation of the context and the response candidates as the input.

\textbf{BERT-VFT}~\cite{whang2020domain}: before fine-tuning, the model also carries out a post-training on training corpora in the same manner as BERT.

\textbf{SA-BERT}~\cite{gu2020speaker}: the model follows BERT-VFT, and further incorporates speaker-aware embeddings.

\begin{table*}[ht!]
    \centering
    \resizebox{0.75\textwidth}{!}{
    \begin{tabular}{l|c|c|c|c|c|c|c}
      \thickhline
      \multirow{2}{*}{\backslashbox{Models}{Metrics}} &   \multicolumn{4}{c|}{\textbf{Ubuntu Corpus}}    &        \multicolumn{3}{c}{\textbf{E-Commerce Corpus}}        \\ \cline{2-8}
      &  R$_2$@1 &  R$_{10}$@1 & R$_{10}$@2 & R$_{10}$@5 & R$_{10}$@1 & R$_{10}$@2  & R$_{10}$@5\\ \hline
      DualLSTM~\cite{lowe-etal-2015-ubuntu} & 0.901 & 0.638 & 0.784 & 0.949 & 0.365 & 0.536 & 0.828\\
      DualLSTM-SL &0.925*&0.724* & 0.858* & 0.969*& 0.518* & 0.722* & 0.933*   \\ \hline
      ESIM~\cite{chen2019sequential}  & 0.950 & 0.796 & 0.874 & 0.975 & 0.570 & 0.767 & 0.948 \\ 
      ESIM-SL & 0.963* & 0.822* & 0.909* & 0.980* & 0.623* & 0.797* & 0.969* \\ 
      \thickhline
    \end{tabular}
    }
    \caption{Evaluation results of two matching models trained with the  proposed self-supervised tasks.  Numbers marked with $*$ mean that the improvement is statistically significant compared with the baseline (t-test with $p$-value $<0.05$).}  
    \label{exp:small-models}
  \end{table*}

\subsection{Implementation Details} 
Following~\citet{gu2020speaker}, we select English uncased $\text{BERT}_\text{base}$ (110M) as the context-response matching model for the Ubuntu dataset and Chinese $\text{BERT}_\text{base}$ model for the E-commerce dataset. We implement the models with the code in \url{https://github.com/huggingface/transformers}.
The maximum lengths of the context and response were set to 448 and 64 as the maximum length of input sequence in BERT is 512.
Intuitively, the last tokens in the context and the previous tokens in the response candidate are more important, so we cut off the previous tokens for the context but do the cut-off in the reverse direction for the response candidate if the sequences are longer than the maximum length.
We choose 32 as the size of mini-batches for training.
On both the Ubuntu dataset and the Douban dataset, we applied domain adaptive post-training before the finetuning procedure following the settings of ~\citet{whang2020domain}.
Training instances of auxiliary tasks are generated dynamically.
We select $\Delta$ (Equation (\ref{eq:cdloss})) in $\{0.2,
0.4,0.6,0.8\}$ and find that $0.6$ is the best choice.
We vary $\alpha$ (Equation (\ref{eq:loss})) in $\{0.1, 0.2, 0.5, 1.0\}$ and choose $\alpha=1.0$ as the trade-off between the learning objectives.
The model is optimized using Adam optimizer with a learning rate set as $3e-5$.
Early stopping on the validation data is adopted as a regularization strategy. 
All the experiment results except ours are cited from previous works.

\subsection{Evaluation Results}
Table~\ref{exp:main-results} reports the results of BERT-SL and all baseline models on the Ubuntu datasets and the E-commerce dataset.
From the evaluation results, we can easily observe that the PLM-based response selection models generally perform better than the models based on various neural architectures. The phenomenon shows the advantage of the pre-trained models on providing strong universal representations for response selection.
Among those PLM-based response selection models, our BERT-SL outperforms the best baseline BERT-VFT in terms of all metrics on both data sets. Specifically, compared to the previous state-of-the-art model, our BERT-SL achieves $2.9$\% absolute improvement in terms of R$_{10}$@1 on the Ubuntu dataset and $4.8$\% absolute improvement on the E-commerce dataset. We conduct statistical tests, and the results indicate that the improvement on all metrics except R$_{10}$@5 on the E-commerce data is statistically significant. The significant improvement demonstrates the effectiveness of our proposed self-supervised learning schema. 
Notably, our method does not increase the inference time compared with existing PLM-based models.

\begin{figure*}[h!]
  \centering
  \subfigure[{Average turns vs. R$_{10}$@1}] { \label{fig:turns}
    \includegraphics[width=0.46\linewidth]{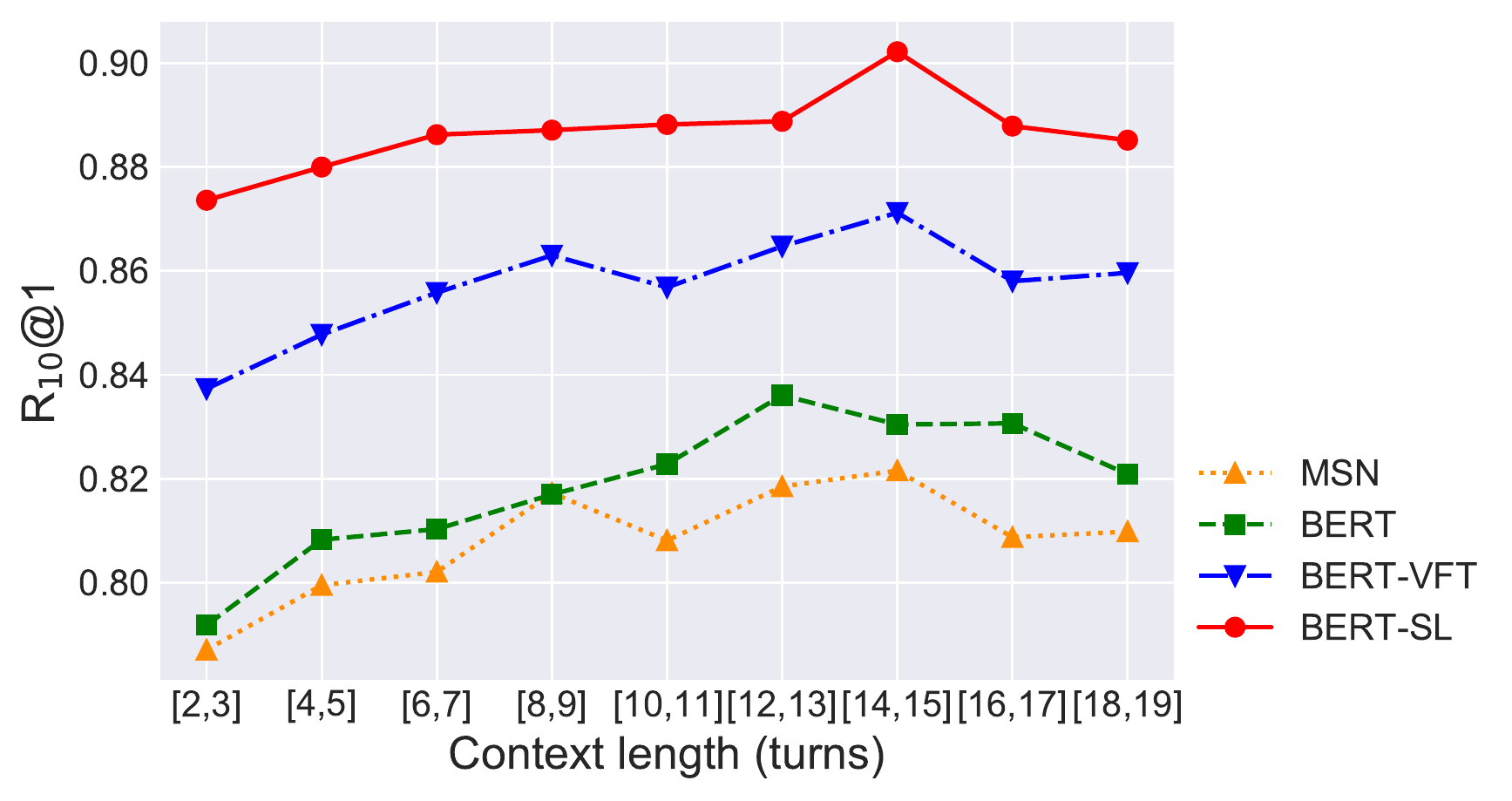}
  }
  \subfigure[{Total context length vs. R$_{10}$@1}] { \label{fig:tokens}
    \includegraphics[width=0.46\linewidth]{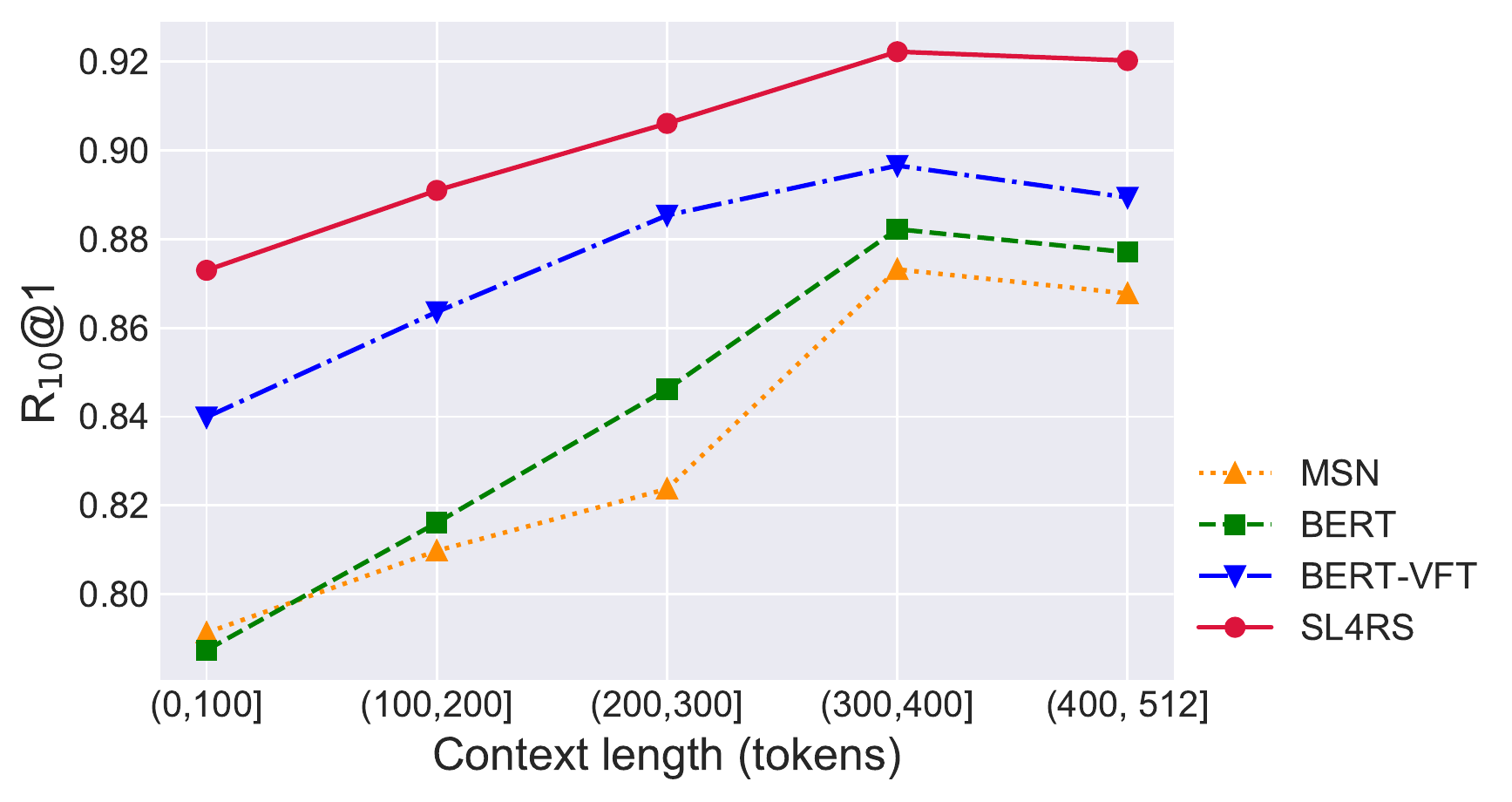}
  }
  \caption{Performance of BERT-SL and its variants across different lengths of contexts. (a) context length is measured by the average number of turns; (b) context length is measured by the total length of the context.}
  \label{fig:context-length}
\end{figure*}

\subsection{Discussions}
\paragraph{Ablation study.}
To investigate the impact of different self-supervised tasks, we conducted a  comprehensive ablation study. 
We keep the architecture of the matching model and remove each self-supervised task individually from the model, and denote the model as ``{BERT-SL} w/o. $\mathcal{T}$", where $\mathcal{T} \in \{\text{NSP, UR, ID, CD}\}$ stand for next session prediction, utterance restoration, incoherence detection and  consistency discrimination respectively.
The detailed results are reported in the last four lines of Table~\ref{exp:main-results}.
First of all, we find that all four self-supervised tasks are useful as removing any of them causes a performance drop on both datasets.
Second, we can conclude that on the Ubuntu data, the rank of the tasks in terms of $R_{10}@1$ is that $\text{ID}>\text{NSP}>\text{CD}>\text{UR}$; and on the E-commerce data, the rank of the tasks is that $\text{CD}>\text{ID}>\text{NSP}>\text{UR}$\footnote{We select $R_{10}@1$ as target metrics in the study of the importance of different tasks because they are more critical than other metrics in real systems of response selection.}.
Among the four tasks, ID plays an important role in improving the response selection task. The reason might be that the ID task can encourage the model to consider the coherence between the context and a response candidate, which acts as complementary to the main task.
It is also noted that removing the utterance restoration task leads to the slightest decrease of the performance 
on both datasets, 
as the feature learned by UR may be redundant with that learned by the token-level mask language modeling in pre-training. Besides, the representation learned by the generative task might have a considerable discrepancy with the discrimination task.
Finally, the CD task is much more important on the E-commerce data than it is on the Ubuntu data, as E-commerce corpora contain more diverse content.

\vspace{1.5mm}
\noindent\textbf{Self-supervised learning for ESIM/DualLSTM.}
We are curious about whether the effectiveness of the proposed self-supervised learning schema depends on the architecture of the response selection model. Therefore, we test our proposed learning schema on some non-PLM-based response selection models, such as dual LSTM~\cite{lowe-etal-2015-ubuntu} and ESIM~\cite{chen2019sequential}. The original two models treat the multi-turn context as a long sequence and are trained with only the context-response task. Thus, we implement two models and jointly train them with the proposed four self-supervised tasks in a multi-task manner.
Table~\ref{exp:small-models} reports the comparison results on both data sets. We observe a consistent and significant improvement of the performance for both DualLSTM and ESIM.
Particularly, with the help of auxiliary self-supervised tasks, a simple ESIM  model  can  even  achieve  better  performance on the Ubuntu dataset  than BERT, which is a bigger pre-trained model.
The results imply that our learning schema is beneficial for various matching architectures, and indicate the effectiveness and generality of the proposed method.

\vspace{1.5mm}
\noindent\textbf{Performance across different lengths of context.}
To analyze how the performance of our proposed BERT-SL varies with different context lengths, we compare BERT-SL with BERT, BERT-VFT and the state-of-the-art non-PLM-based response selection models (a.k.a. MSN).
In this work, context length is measured by (1) number of turns and (2) number of all tokens in a context.
Figure~\ref{fig:context-length} shows how the performance of the four models varies across contexts with different lengths.
We can observe that the performance of all models first increases monotonically when the context length increases, and then fluctuates or even drops when context length keeps increasing.
The reason might be that when only a few utterances are available in the context, the model could not capture enough information for matching, but when the context becomes long enough, noises will be brought to matching due to the topic shift in dialogue.
Across the different lengths of the context, our $\text{BERT-SL}$ can always achieve better performance than $\text{BERT-VFT}$ as well as other baselines.
It is worth noting that the performance of our $\text{BERT-SL}$ is more stable than other models across different turns of the context, and drops more slightly than other models for a long context.
The results imply that our learning schema improves the capability of the matching model to deal with long contexts or short context.

\vspace{-1.5mm}
\section{Related Works}
With the advance of natural language processing, building an intelligent dialogue system with data-driven approaches~\cite{vinyals2015neural,lowe-etal-2015-ubuntu} has drawn increasing interests in recent years.
Most existing approaches are either generation-based or retrieval-based.
The former synthesize a response word by word via natural language generation techniques~\cite{vinyals2015neural,serban2016building},
while the latter
select the most appropriate response from a set of candidates~\cite{wang-etal-2013-dataset,wu-etal-2017-sequential,whang2020domain}.
We focus on retrieval-based methods in this paper.
Earlier studies pay attention to constructing single-turn context-response matching models where only a single utterance is considered or multiple utterances in the context are concatenated into a long sequence for response selection~\cite{wang-etal-2013-dataset,hu2014convolutional,lowe-etal-2015-ubuntu}.
Recently, most studies focus on the multi-turn scenario where each utterance in the context first interacts with the response candidate, and then the matching features are aggregated according to the sequential dependencies of the multi-turn context~\cite{zhou-etal-2016-multi,wu-etal-2017-sequential,yan2016learning,zhou2018multi,tao-etal-2019-one}, and they usually adopt the \emph{representation-matching-aggregation} paradigm to build the matching models. 
Following the paradigm, ~\citet{tao2019multi} and \citet{wang2019multi} further consider multiple granularities of representations for matching. Besides, to tackle the side effect of using too much context, ~\citet{yuan-etal-2019-multi} utilizes a multi-hop selector to select the relevant utterances in the context for response matching.

Recently, pre-trained language models~\cite{devlin-etal-2019-bert,yang2019xlnet,liu2019roberta} have shown impressive benefits for various downstream NLP tasks, and some researchers tried to apply them on response selection. 
~\citet{vig2019comparison} utilizes BERT to represent each utterance-response pair and aggregate these representations to calculate the matching score.
~\citet{whang2020domain} treat the context as a long sequence and perform context-response matching with the BERT. Besides, the model also introduces the next utterance prediction and mask language modeling tasks borrowed from BERT during the post-training on dialogue
corpus to incorporate in-domain knowledge for the matching model.
Following ~\citet{whang2020domain}, ~\citet{gu2020speaker} propose to heuristically incorporate speaker-aware embeddings into BERT to promote the capability of context understanding in multi-turn dialogues.

Self-supervised learning has become a significant direction in the AI community and has contributed to the success of pre-trained language models~\cite{devlin-etal-2019-bert,yang2019xlnet,liu2019roberta}.
Inspired by this, some researchers propose to learn down-stream tasks with auxiliary self-supervised tasks.
In this manner, models can effectively learn task-related knowledge with a fixed amount of training data and produce better features for the primary task.
Existing works have explored self-supervised tasks in text classification~\cite{yu-jiang-2016-learning}, summarization~\cite{wang-etal-2019-self} and response generation~\cite{zhang2019consistent,zhao2020learning}.
The work is unique in that we design several self-supervised tasks according to the characteristics of the dialogue data to improve the multi-turn response selection and our learning schema can bring consistent and significant improvement for both traditional context-response matching models and large-scale pre-trained language models.

\section{Conclusion}
In this paper, we propose learning a context-response matching model with four auxiliary self-supervised tasks designed for the dialogue data.
Jointly trained with these auxiliary tasks, the matching model can effectively learn task-related knowledge contained in dialogue data, achieve a better local optimum and produce better features for response selection.
Experiment results on two benchmarks indicate that the proposed auxiliary self-supervised tasks bring significant improvement for multi-turn response selection in retrieval-based dialogues, and our PLM-based model achieves new state-of-the-art results on both datasets.

\nocite{langley00}
\bibliography{main}
\bibliographystyle{icml2020}

\end{document}